\documentclass[letterpaper, 10 pt, conference]{ieeeconf}  

\IEEEoverridecommandlockouts                              

\usepackage[l2tabu,orthodox]{nag}

\usepackage[
backend=biber, 
bibencoding=utf8,
style=ieee, 
sorting=none, 
natbib=true, 
doi=false, 
isbn=false, 
url=false, 
eprint=false, 
maxcitenames=1, 
mincitenames=1,
maxbibnames=9,
]{biblatex}

\usepackage[draft,pdftex,colorlinks]{hyperref}

\usepackage[printonlyused]{acronym}

\usepackage{siunitx}
\usepackage[all]{nowidow}

\usepackage[dvipsnames]{xcolor}

\usepackage{lipsum}

\usepackage[pdftex]{graphicx}

\usepackage{epstopdf}

\usepackage{import}

\graphicspath{{./latexGoodPractices/}}

\usepackage{booktabs}

\usepackage{tabu}

\usepackage{amssymb,amsfonts,amsmath,amscd}

\usepackage{bm} 

\newcommand{\bbm}{\begin{bmatrix}}
\newcommand{\ebm}{\end{bmatrix}}

\usepackage[normalem]{ulem} 

\makeatletter
\usepackage[utf8]{inputenc}
\usepackage{commath}
\usepackage{mathtools}
\usepackage{mathabx}
\usepackage{siunitx}
\usepackage[ruled,vlined]{algorithm2e}
\usepackage{caption}
\usepackage{graphicx}
\hypersetup{final}

\let\oldtheequation\theequation
\renewcommand\tagform@[1]{\maketag@@@{\ignorespaces#1\unskip\@@italiccorr}}
\renewcommand\theequation{(\oldtheequation)}
\makeatother

\graphicspath{{media/}}

\acrodef{lidar}{Light Detection And Ranging}
\acrodefplural{lidar}{Light Detection And Ranging}
\acrodef{ICP}{Iterative Closest Point}
\acrodef{GNSS}{Global Navigation Satellite System}
\acrodef{IMU}{Inertial Measurement Unit}
\acrodef{DOF}{Degrees Of Freedom}
\acrodef{RTK}{Real Time Kinematics}
\acrodef{GPS}{Global Positioning System}
\acrodef{SVD}{Singular Value Decomposition}
\acrodef{UAV}{Unmanned Aerial Vehicle}
\acrodef{IIR}{Infinite Impulse Response}
\acrodef{CAD}{Computer-aided design}
\acrodef{SLAM}{Simultaneous Localization and Mapping}
\acrodef{DARPA}{Defense Advanced Research Projects Agency}

\title{\LARGE{\textbf{
Gravity-constrained point cloud registration}}
}

\author{Vladim\'ir Kubelka, Maxime Vaidis and François Pomerleau$^{1}$
\thanks{*This research was supported by the  Natural  Sciences and Engineering  Research  Council of  Canada  (NSERC)  through the grant CRDPJ 527642-18 SNOW (Self-driving Navigation Optimized for Winter).}
\thanks{$^{1}$The authors are with Northern Robotics Laboratory, Université Laval, Québec City, Canada,
  {\tt{\small{$\{$vladimir.kubelka, maxime.vaidis, francois.pomerleau$\}$ @norlab.ulaval.ca}}}}%
}

\begin{document}

\maketitle
\thispagestyle{empty}
\pagestyle{empty}

\begin{abstract}
Visual and lidar \ac{SLAM} algorithms benefit from the \ac{IMU} modality.
The high-rate inertial data complement the other lower-rate modalities.
Moreover, in the absence of constant acceleration, the gravity vector makes two attitude angles out of three observable in the global coordinate frame.
In visual odometry, this is already being used to reduce the 6-\ac{DOF} pose estimation problem to 4-DOF.
In lidar \ac{SLAM}, the gravity measurements are often used as a penalty in the back-end global map optimization to prevent map deformations.
In this work, we propose an \ac{ICP}-based front-end which exploits the observable \ac{DOF} and provides pose estimates aligned with the gravity vector.
We believe that this front-end has the potential to support the loop closure identification, thus speeding up convergences of global map optimizations.
The presented approach has been extensively tested in large-scale outdoor environments as well as in the Subterranean Challenge organized by \ac{DARPA}.
We show that it can reduce the localization drift by 30\,\% when compared to the standard 6-\ac{DOF} \ac{ICP}.
Moreover, the code is readily available to the community as a part of the libpointmatcher library.
\end{abstract}

\acresetall

\section{Introduction}
Among the \ac{SLAM} approaches for mobile robots, cameras and \ac{lidar} modalities are still the dominant front-end choices~\cite{Cadena2016}.
Modern visual \ac{SLAM} and visual odometry solutions benefit from the complementary nature of \ac{IMU} modalities, which provide high-rate linear acceleration and rotational velocity between captured camera frames~\cite{Forster2017}.
Moreover, the acceleration sensed by \acp{IMU} allows the detection of the gravity vector. 
This vector can be used to observe two attitude angles out of three in a global coordinate frame, thus reducing the visual odometry pose estimation problem from 6-\ac{DOF} to 4-\ac{DOF}~\cite{Zhang2018}.
Similarly, in the domain of \ac{lidar} \ac{SLAM}, using \acp{IMU} is beneficial for precise point cloud alignment \cite{Shan2020}.
The two observable attitude angles are often used as penalties in \ac{SLAM} back-ends during the global map optimization step.
These penalties have the benefit of aligning the map with the gravity vector~\cite{Zlot2014, Ye2019, Agha2021nebula}.
This technique reduces map deformations, especially in large environments with a limited number of possible loop closures, such as in underground mines~\cite{Zlot2014}.
We will refer to the use of gravity vector in a \ac{SLAM} back-end as a loosely-coupled solution.

\begin{figure}[htbp]
    \centering
    \includegraphics[width=\linewidth]{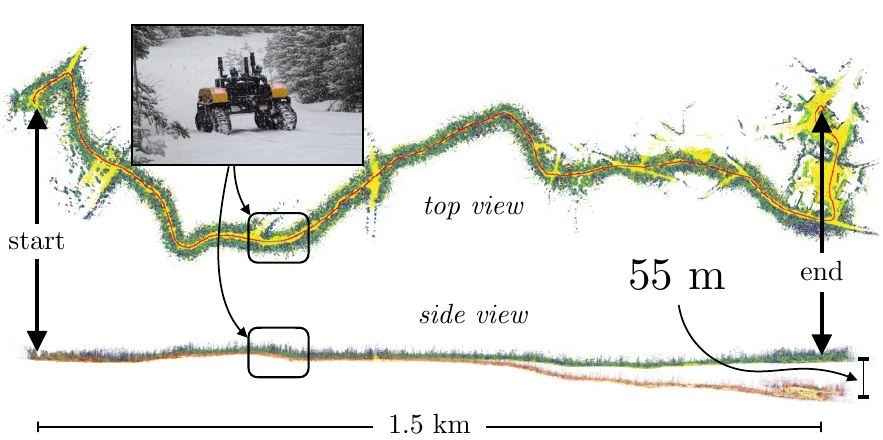}
    \caption{
    Example of a drifting localization-based solely on a front-end relying on lidar registration.
    The top view shows the 3D reconstruction of the forest path over \SI{1.5}{\km}.
    Color is used to highlight the structure of the environment, with yellow representing the ground.
    The side view of the same path shows reconstruction drifting by \SI{55}{\m} in elevation with red points and a gravity constrained map in green.
    }
    \label{fig:drift}
\end{figure}

\ac{SLAM} front-end solutions relying on \acp{lidar} typically build an incremental 3D map of the environment by registering a current scan using a variant of the original \ac{ICP} algorithm~\cite{Besl1992}.
The \ac{ICP} algorithm estimates a rigid transformation by iteratively finding corresponding points between two point clouds and minimizing an alignment error.
This solution ensures local crisp maps, yet inevitably suffers from global drift \cite{Pomerleau2013}.
This phenomenon is shown in \autoref{fig:drift}, where a robot navigated autonomously on a \SI{1.5}{\km} path crossing a forest, while only localizing on subsequent scans.
This drift problem can be mitigated by \ac{SLAM} back-ends applying pose graph optimization \citep{Thrun2006,Cadena2016}. 
Global cues, such as loop closures, are used as constraints to optimize all estimated transformations between the individual lidar scans to ensure global consistency.
\ac{SLAM} back-ends rely on pose-graph optimizers, which require uncertainty estimations in the form of covariance matrices for all transformations.
Unfortunately, when it comes to adding constraints produced by \ac{ICP}, it was shown that, although the uncertainty can be learned or sampled, the Gaussian distribution assumption does not hold up well in complex 3D environments~\citep{Landry2019}.

In this paper, we present a tightly-coupled variant of the \ac{ICP} algorithm, which exploits the directly observable \acp{DOF} in the \ac{IMU} orientation.
This way, we obtain pose estimates aligned with the gravity vector.
The proposed modification is intended as a \ac{lidar} \ac{SLAM} front-end that limits mapping drift.
We believe that this front-end has the potential to support the loop closure identification, thus speeding up convergences of global map optimizations.  
Our solution has been extensively tested in large-scale outdoor environments as well as in the \ac{DARPA} Subterranean Challenge (SubT).\footnote{\url{https://www.darpa.mil/program/darpa-subterranean-challenge}}
Moreover, the code is readily available to the community as a part of the \texttt{libpointmatcher} library.\footnote{\url{https://github.com/ethz-asl/libpointmatcher}}

The remainder of the paper is structured as follows: \autoref{sec:related_work} provides a brief overview of the \ac{SLAM} algorithms that benefit from the \ac{IMU} modality. \autoref{sec:theory} introduces the modified 4-DOF \ac{ICP} algorithm including the derivation of the internal optimization step. The following \autoref{sec:experiments} presents our experimental setup and \autoref{sec:results} discusses the experiment results. Finally, \autoref{sec:conclusion} summarizes our findings and concludes the paper.

\section{Related work}
\label{sec:related_work}


In this section, we first briefly present the visual-inertial SLAM and odometry solutions.
Then, the lidar-based systems follow.
We divide them into two subgroups according to the manner the IMU modality is exploited.
The first subgroup consists of solutions that pre-integrate inertial data into pose increments and use them as penalties in state estimation.
The second subgroup is specific by exploiting the gravity force vector that serves as an additional global attitude measurement.
Finally, we put our topic, lidar registration, in context with these works.

In the field of visual odometry, the works of \citet{Kelly2011} and \citet{Jones2011} presented filtering methods for motion estimation from monocular camera supported by inertial measurements.
Besides pose estimation, the algorithms allowed online self-calibration of the camera and \ac{IMU} coordinate frames.
The following extensive research in the field gradually evolved towards graph-based optimization methods, which became computationally feasible thanks to incremental smoothing~\citep{Kaess2011}.
To include inertial information into the factor graphs, \citet{Forster2017} improved the \ac{IMU} preintegration technique, originally proposed by \citet{Lupton2012}.
The preintegration allows condensing high-rate inertial measurements into compact inter-frame motion penalties.
These, in turn, allow real-time operation of visual-inertial odometry and \ac{SLAM}.
When the \ac{IMU} is not subject to a constantly accelerated motion, the sensed gravitational force makes two of the six pose \ac{DOF} observable.
As stated in \citet{Zhang2018}, in a visual-inertial odometry framework, the gravity vector simplifies the problem to a 4-\ac{DOF} pose estimation.
These \acp{DOF} are three positions and one rotation around the vertical axis in a gravity-aligned global reference frame.
In our work, we leverage these observations and propose a solution to include this very principle in the lidar-scan-matching process instead of camera-based frameworks.

In the field of \ac{lidar} \ac{SLAM}, the \ac{IMU} modality has attracted attention as well.
\citet{Shan2020} tightly integrated \ac{IMU}, \ac{lidar} and \ac{GPS} in a factor graph optimized by the iSAM2 \cite{Kaess2011} algorithm.
The \ac{GPS} and the ability to identify loop closures lead to globally consistent metric maps.
A lightweight lidar-inertial ego-motion estimator was presented by \citet{Qin2020}.
They used an iterated error-state Kalman filter to achieve real-time performance, while still gaining from the benefits of tight integration of the sensory data.
Similarly, a solution for an \ac{UAV} that integrated \ac{lidar}, \ac{IMU} and ultra-wide-band ranging modalities was proposed by \citet{Nguyen2021}.
Each modality provided penalties for a sliding-window optimization carried out by the Ceres solver.\footnote{\url{http://ceres-solver.org}}
We see that these works treat the pre-integrated IMU pose increments as penalties in the state estimation process.
Closer to our application, \citet{Palieri2021} introduced their lidar odometry system used in the DARPA SubT Challenge.
The \ac{IMU} modality is integrated more loosely, but it is important for removing motion distortion from lidar scans.
They also stressed the necessity of robust operation in harsh conditions and their design dynamically handles sensor modality failures.

Representing the second subgroup, the following works explicitly use the gravity force measurement for stabilizing the vertical orientation of their world coordinate frame and thus their map representation.
\citet{Zlot2014} proposed a lidar mapping solution for underground mines.
Apart from benefiting from the inertial measurements as a prior for matching point clouds, they explicitly used gravity vector constraints to ensure the vertical orientation.
The gravity vector is one of several linear constraints in their optimization process that smooths prior trajectory. 
In a tightly integrated lidar-inertial \ac{SLAM} system for autonomous vehicles, \citet{Ye2019} used rotation constraints in the global optimization step.
The team COSTAR from the SubT competition added explicit gravity factors in their graph representation used in global optimization \cite{Agha2021nebula}.
In our work, we benefit from the two observable orientation \acp{DOF} that also rise from the gravity measurement.

Our solution focuses on the front-end of \ac{SLAM} systems relying principally on \ac{lidar} (i.e., on the point-cloud registration).
The \ac{ICP} algorithm iteratively finds corresponding points between two point clouds and looks for a rigid transformation minimizing the alignment error.
\ac{ICP} ensures local consistency, yet inevitably suffers from global drift \cite{Pomerleau2013}.
This drift can be mitigated by \ac{SLAM} techniques of global map optimization based on spatial (e.g., loop closures) and measurement constraints (e.g., \ac{GPS} measurements) \cite{Cadena2016}.
We propose to modify directly the \ac{ICP} minimization process to exploit observable \acp{DOF} from the gravity vector.
This leads to searching for pose estimates in a 4-\ac{DOF} space while limiting the vertical and axial drifts of the local pose estimates.

\section{Theory}
\label{sec:theory}
This section first provides a high-level overview of the mapping framework that we use to evaluate the proposed \ac{ICP} modification.
Then, it derives the  4-\ac{DOF} minimization equations for \ac{ICP} itself. 

\subsection{ICP Mapper}

The mapping and subsequent localization are achieved using the freely available \texttt{norlab\_icp\_mapper}.\footnote{\url{https://github.com/norlab-ulaval/norlab\_icp\_mapper}}
The mapper is lightweight and essentially wraps the \ac{ICP} functionality while performing input lidar data filtering and memory management for the point cloud map.
No global optimization is performed in our setting, the point cloud map grows incrementally as the robot navigates through the environment.
The mapper performs the following steps: 
\begin{enumerate}
    \item Transform the input lidar scan into the world frame according to the initial pose estimate $\widecheck{\bm{T}}$
    \item Register the scan with the map using the \ac{ICP} 
    \item Insert the scan inside the map
\end{enumerate}
The initial estimate $\widecheck{\bm{T}}$ is composed of a translation vector based on the robot odometry and of an orientation quaternion based on the \ac{IMU}.
The input lidar scans are randomly sub-sampled to maintain real-time performance on mobile robots.
The input scans are also corrected for the motion distortion according to the odometry and \ac{IMU} data.
Finally, only a subset of the aligned scan points is merged into the map to keep its density bounded.

\subsection{4-DOF ICP}

The \ac{ICP} algorithm aims at estimating a rigid transformation $\widehat{\bm{T}}$ that best aligns a set of 3D points $\mathcal{Q}$ (i.e., a \emph{map} point cloud) with a second set of 3D points $\mathcal{P}$ (i.e., \emph{scan} point cloud), by minimizing the error function $e$:
\begin{equation}
\label{eq:icp_general}
\widehat{\bm{T}} = \arg \min_{\bm{T}} \: e \left( \mathcal{Q}, \mathcal{P} \right).
\end{equation}
In our approach, we choose the \emph{point-to-plane} definition of $e$ \cite{Chen1992} since it outperforms the \emph{point-to-point} definition in most cases \cite{Pomerleau2013}.
However, the 4-DOF modification can be implicitly extended to the \emph{point-to-Gaussian} error definition since the latter can be translated to the \emph{point-to-plane} as shown by \citet{Babin2019a}.
We first define our transformation parameter $\bm{T}$ as a 4D vector $\bm{\tau}$:
\begin{equation}
\bm{\tau} =
\left[ \begin{array}{c}
\gamma \\
\bm{t}
\end{array} \right]  =
\left[ \begin{array}{c}
\gamma \\
t_x \\
t_y \\
t_z
\end{array} \right],
\end{equation}
where $\gamma$ is the rotational component along the \textbf{z} axis, and $t_x$, $t_y$ and $t_z$ are the translation components.
Compared to the classical 6-DOF definition of $e$, please note that $\bm{\tau}$ misses the rotational components along the \textbf{x} and \textbf{y} axes. 
We then follow with the standard definition of the point-to-plane error function
\begin{equation} \label{eq:p2plane}
e = \sum\limits_{k=1}^K \left\| \left[(\bm{R}\bm{p}_k + \bm{t})- \bm{q}_k\right] \cdot \bm{n}_k \right\|_2 ,
\end{equation}
where $\bm{n}_k$ is the normal vector representing the surface at the point $\bm{q}_k$, $\bm{R}=R(\gamma)$ is a rotation matrix and the index $k$ represents paired points from $\mathcal{Q}$ and $\mathcal{P}$.
Closely following the minimization derivation from \citet{Pomerleau2015a}, yet modified for 4-DOF, the rotation matrix $\bm{R}$ performs only the yaw rotation.
The other two rotations are not necessary since the $\mathcal{P}$ point cloud had already been pre-aligned by the mapper according to the initial $\widecheck{\bm{T}}$ pose estimate.
The minimization method relies on rotation matrix linearization.
This linearization can be achieved using the small-angle approximation:
\begin{equation} \label{eq:rotLin}
\bm{R} = R(\gamma)  
\approx \underbrace{\left[ \begin{array}{ccc} 
0 & -1 & 0 \\
1 & 0 & 0 \\
0 & 0 & 0
\end{array} \right]}_{\bm{\Gamma}} \gamma + \bm{I} ,
\end{equation}
where $\bm{I}$ is a unit matrix.
In the context of ICP, the impact of linearization is reduced through the iterative process of the whole registration algorithm.
Combining \ref{eq:p2plane} with \ref{eq:rotLin}, we can approximate the objective function as
\begin{align*}
e &\approx \sum\limits_{k=1}^K \left\| [(\bm{\Gamma}\gamma + \bm{I})\bm{p}_k + \bm{t} - \bm{q}_k] \cdot \bm{n}_k \right\|_2 \\
 &\approx \sum\limits_{k=1}^K \left\| \gamma (\bm{\Gamma} \bm{p}_k) \cdot \bm{n}_k + \bm{p}_k \cdot \bm{n}_k + \bm{t} \cdot \bm{n}_k - \bm{q}_k \cdot \bm{n}_k \right\|_2 ,
\end{align*}
which can be rewritten using the \emph{scalar triple product} and by reorganizing the terms, such that
\begin{align*}
e &\approx \sum\limits_{k=1}^K \left\| \gamma  \underbrace{(\bm{\Gamma} \bm{p}_k) \cdot \bm{n}_k}_{{\text{\textit{\c{c}}}_k}} + \bm{t} \cdot \bm{n}_k - \underbrace{(\bm{q}_k - \bm{p}_k)}_{\bm{d}_k} \cdot \bm{n}_k  \right\|_2 \\
 &\approx \sum\limits_{k=1}^K \left\| \gamma \, \text{\textit{\c{c}}}_k + \bm{t} \cdot \bm{n}_k - \bm{d}_k \cdot \bm{n}_k  \right\|_2 .
\end{align*}
We can then minimize the error $e$ with respect to $\bm{r}$ and $\bm{t}$ and setting the partial derivatives to zero
\begin{align*}
\frac{\partial e}{\partial \gamma} &= \sum\limits_{k=1}^K 2 \text{\textit{\c{c}}}_k (\gamma \, \text{\textit{\c{c}}}_k + \bm{t} \cdot \bm{n}_k - \bm{d}_k \cdot \bm{n}_k) = \bm{0} \\
\frac{\partial e}{\partial \bm{t}} &= \sum\limits_{k=1}^K 2 \bm{n}_k (\gamma \, \text{\textit{\c{c}}}_k + \bm{t} \cdot \bm{n}_k - \bm{d}_k \cdot \bm{n}_k) = \bm{0} .
\end{align*}
We can assemble these derivatives under the linear form $\bm{A}\bm{\tau}=\bm{b}$, by bringing the independent variables on the right side of the equation
\begin{align*}
\sum\limits_{k=1}^K 
\left[ \begin{array}{cc}
\text{\textit{\c{c}}}_k (\gamma \text{\textit{\c{c}}}_k) + \text{\textit{\c{c}}}_k (\bm{t} \cdot \bm{n}_k)   \\
\bm{n}_k (\gamma \text{\textit{\c{c}}}_k) + \bm{n}_k (\bm{t} \cdot \bm{n}_k)  
\end{array} \right]
&=
\sum\limits_{k=1}^K 
\left[ \begin{array}{cc}
\text{\textit{\c{c}}}_k (\bm{d}_k \cdot \bm{n}_k)   \\
\bm{n}_k (\bm{d}_k \cdot \bm{n}_k)
\end{array} \right]
\\
\sum\limits_{k=1}^K 
\left[ \begin{array}{cc}
\text{\textit{\c{c}}}_k^2 \gamma + \text{\textit{\c{c}}}_k \bm{n}_k^\top \bm{t}    \\
\bm{n}_k \text{\textit{\c{c}}}_k \gamma + \bm{n}_k \bm{n}_k^\top \bm{t}  
\end{array} \right]
&=
\sum\limits_{k=1}^K 
\left[ \begin{array}{cc}
\text{\textit{\c{c}}}_k (\bm{d}_k \cdot \bm{n}_k)   \\
\bm{n}_k (\bm{d}_k \cdot \bm{n}_k)
\end{array} \right]
\\
\sum\limits_{k=1}^K 
\left[ \begin{array}{cc}
\text{\textit{\c{c}}}_k^2 & \text{\textit{\c{c}}}_k \bm{n}_k^\top    \\
\text{\textit{\c{c}}}_k \bm{n}_k  & \bm{n}_k \bm{n}_k^\top   
\end{array} \right]
\left[ \begin{array}{cc}
\gamma \\
\bm{t}
\end{array} \right]
&=
\sum\limits_{k=1}^K 
\left[ \begin{array}{cc}
\text{\textit{\c{c}}}_k \\
\bm{n}_k 
\end{array} \right] (\bm{d}_k \cdot \bm{n}_k) ,
\end{align*}
which brings us to the linear system of equations
\begin{align} \label{eq:minPointToPlaneBis}
\underbrace{
\sum\limits_{k=1}^K 
\left[ \begin{array}{cc}
\text{\textit{\c{c}}}_k   \\
\bm{n}_k  
\end{array} \right]
\left[ \begin{array}{cc}
\text{\textit{\c{c}}}_k & \bm{n}_k^\top \\
\end{array} \right]
}_{\bm{A}_{4 \times 4}}
\bm{\tau}
 = 
\underbrace{
\sum\limits_{k=1}^K
\left[ \begin{array}{c}
\text{\textit{\c{c}}}_k \\
\bm{n}_k \\
\end{array} \right] 
(\bm{d}_k \cdot \bm{n}_k)
}_{\bm{b}_{4 \times 1}} .
\end{align}
Once the matrix $\bm{A}$ and the vector $\bm{b}$ can be constructed, the linear system of \autoref{eq:minPointToPlaneBis} can be resolved for $\bm{\tau}$ using the Cholesky decomposition.
Implementing such solution will require a loop for the summations over $K$ to build $\bm{A}$ and $\bm{b}$.
An alternative formulation relying on dense matrix multiplication can be computed by assembling
\begin{equation*}
\bm{G}=
\underbrace{
\left[\cdots\begin{array}{c}(\bm{\Gamma} \bm{p}_k) \cdot \bm{n}_k\\\bm{n}_k\end{array}\cdots\right]
}_{4 \times K}, \;\;\;
\bm{h}=
\underbrace{
\left[\begin{array}{c}\vdots\\(\bm{q}_k-\bm{p}_k)\cdot\bm{n}_k\\\vdots\end{array}\right]
}_{K \times 1} ,
\end{equation*}
leading to 
\begin{equation}
\bm{A}\bm{\tau} = \bm{b} 
\;\;\; \Leftrightarrow \;\;\;
\bm{G}\bm{G}^\top\bm{\tau} = \bm{G}\bm{h} , \label{eq:minPointToPlaneCompact}
\end{equation}
which is the final form we use in our implementation.
The \ac{ICP} algorithm iteratively adjusts $\bm{\tau}$ according to \ref{eq:minPointToPlaneCompact} until the convergence of $e$, which indicates the best achievable alignment.
Since our definition of $\bm{\tau}$ does not involve the rotations in the \textbf{x}-\textbf{y} plane, the resulting alignment and consequently the estimated pose follows the initial \ac{IMU} values (i.e., the roll and pitch angles remain the same.)

\section{Experiments}
\label{sec:experiments}
Firstly, the 4-DOF point-to-plane minimizer is readily available in the \texttt{libpointmatcher} library under the \texttt{force4DOF} option and can be tested with the \texttt{norlab\_icp\_mapper}.\footnote{\url{https://github.com/norlab-ulaval/norlab_icp_mapper}}

In this section, we present two sets of experiments that demonstrate the performance of the 4-DOF \ac{ICP} variant and the following \autoref{sec:results} then discusses the achieved results.
The first set of experiments consists of two runs recorded during the post-event testing session at the DARPA SubT finals (see \autoref{fig:darpa_context}).
The second set of experiments covers a \SI{1.5}{km} long forest trail in the Montmorency research forest.\footnote{\url{https://www.foretmontmorency.ca/en/}}
A large tracked robot repeated this path five times in both directions during several days in March 2021.

\subsection{DARPA SubT underground environment}

\begin{figure}[htbp]
    \centering
    \includegraphics[ width=0.99\linewidth]{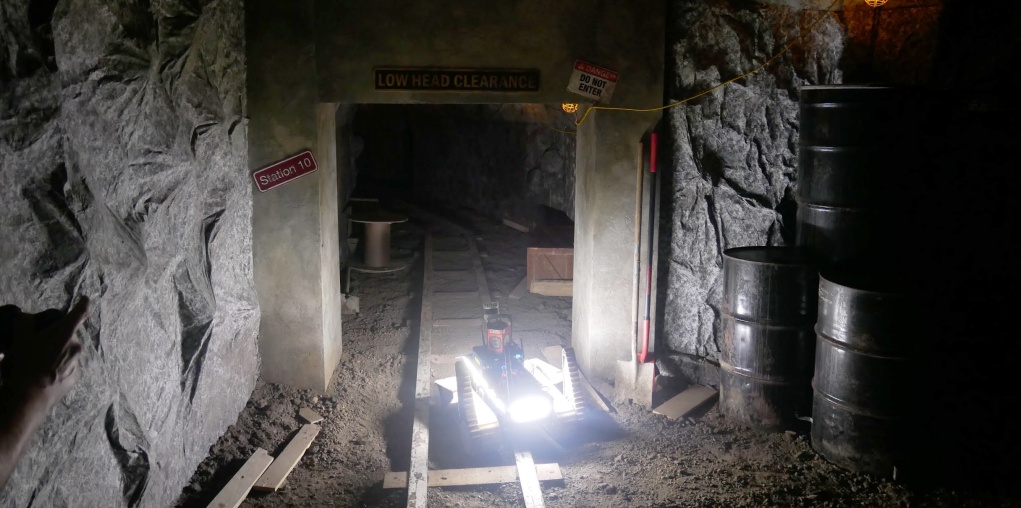}
    \caption{
    The tracked robot driving through the Mine underground environment. It is equipped with an Ouster OS0 128-beam lidar and an Xsens MTI-30 \ac{IMU}.}
    \label{fig:darpa_robot}
\end{figure}

\begin{figure}[htbp]
    \centering
    \includegraphics[trim=1200 0 1100 0,clip, width=\linewidth]{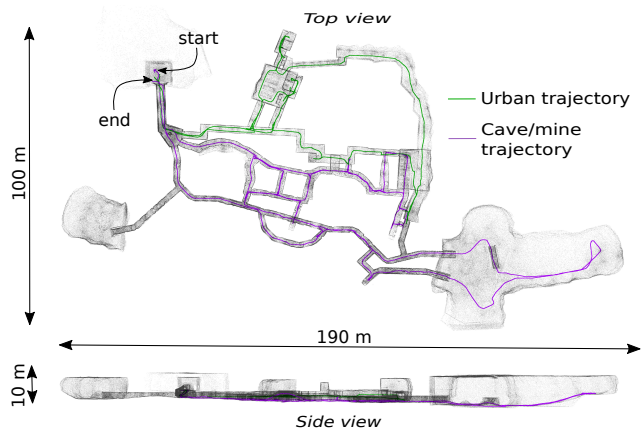}
    \caption{
    The underground environment of the final SubT round. The green and violet trajectories denote the \emph{Urban} and \emph{Mine/Cave} experiments recorded during the post-event testing.
    }
    \label{fig:darpa_context}
\end{figure}

Our laboratory, as a member of the CTU-CRAS-NORLAB team, deployed the 4-DOF \ac{ICP} variant in the Urban circuit and during the competition finals.
Its necessity became apparent after the first Mine circuit, where the mapping and localization drifted in a long entrance tunnel.
This drift would jeopardize the localization of the scored artifacts in the competition. 
In this publication, we present two experiments recorded after the competition finals in Kentucky, during the post-event testing.
The robot shown in \autoref{fig:darpa_robot} was driven through the underground environment in two separate runs, once through the Urban section (green trajectory in \autoref{fig:darpa_context}) and then through the Mine and Cave sections (purple trajectory).
We recorded all sensory data and we are thus able to compare the \ac{ICP} variants offline.
Nevertheless, the mapper performs the same way when launched directly on the robot in real-time.
Furthermore, the competition organizers provide a precise, point-cloud map of the environment.
It allows us to compute ground-truth reference poses with uncertainty comparable to the \ac{lidar} sensor noise.
For this purpose, we initialize the mapper with the reference map and disable the mapping functionality, thus letting the mapper localize within the reference map.

\subsection{Montmorency forest large-scale outdoor environment}

\begin{figure}
    \centering
        \centering
        \includegraphics[width=0.99\linewidth]{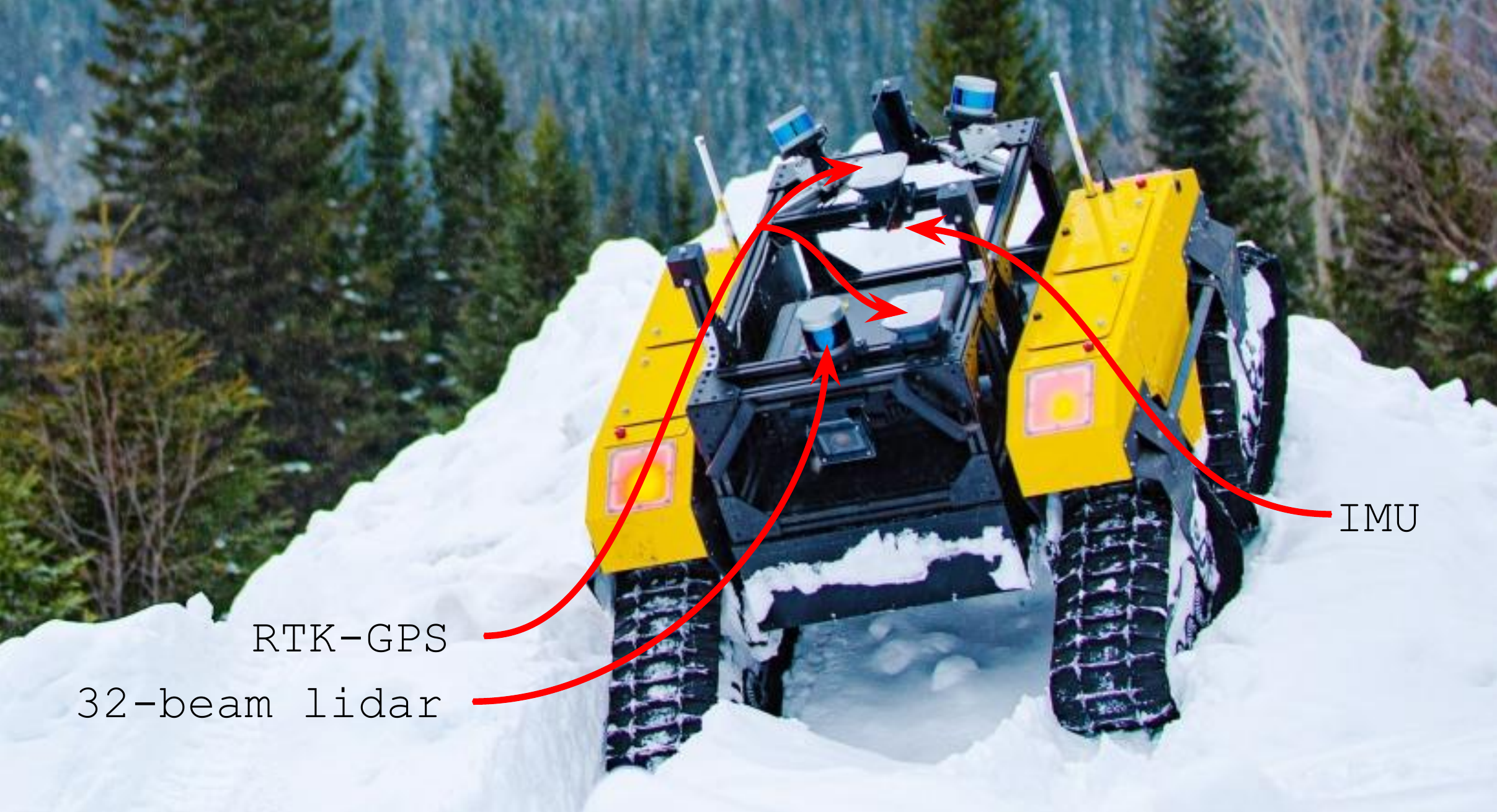}
        \caption{The Clearpath Warthog tracked robot. The navigation is based on the front 32-beam RS-LiDAR-32, Xsens MTI-100 IMU and wheel odometry. The robot is also equipped with two Emlid Reach RS+ RTK-GPS receivers used for reference positioning. The other installed sensors were not used for the experiment.}
        \label{fig:forest_warthog}
\end{figure}

\begin{figure}
    \centering
        \centering
        \includegraphics[width=0.99\linewidth]{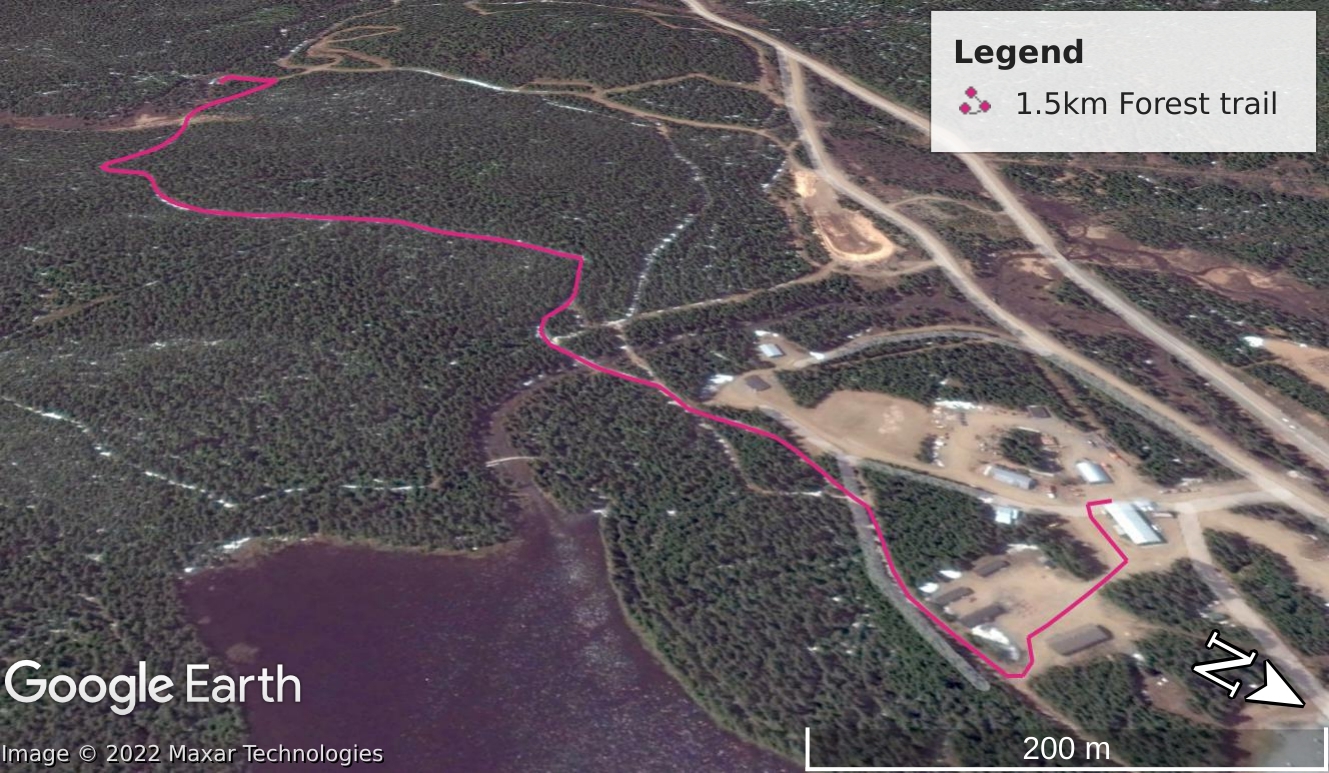}
    \caption{The $1.5\,$km forest trail in the Montmorency forest autonomously navigated in both directions. The map was adapted from \copyright Google Earth.}
    \label{fig:forest_google}
\end{figure}

During the winter season of 2021, we deployed a large tracked Clearpath Warthog robot (\autoref{fig:forest_warthog}) in the research forest.
The purpose of this expedition was the development and testing of a Teach\&Repeat functionality for autonomous outdoor navigation (the detailed report \cite{Baril2021} is in a review process at the moment).
From the data recorded there, we choose a \SI{1.5}{\km} forest trail to demonstrate the impact of our approach to mapping and localization in the large-scale outdoor environment of a sub-arctic forest.
The robot passed the trail shown in \autoref{fig:forest_google} five times in separate experiments during three days.
We tested both directions and the weather conditions varied between sub-zero snowy weather and above-zero rain, therefore altering the character of the snow cover.
Thanks to accurate reference positioning by RTK-GPS receivers on the robot, we can provide the ground truth positioning and thus evaluate the localization error.

\section{Results}
\label{sec:results}
We evaluate the effect of the 4-DOF \ac{ICP} on localization accuracy by comparing the mapper output with the available ground truth.
The metric we choose is the Absolute Trajectory Error (ATE) following \citet{Sturm2012}.
We show its \textbf{z} component and also its modulus.

\subsection{DARPA SubT underground environment}

\begin{figure}[htbp]
    \centering
    \includegraphics[trim=0 0 0 0,clip, width=\linewidth]{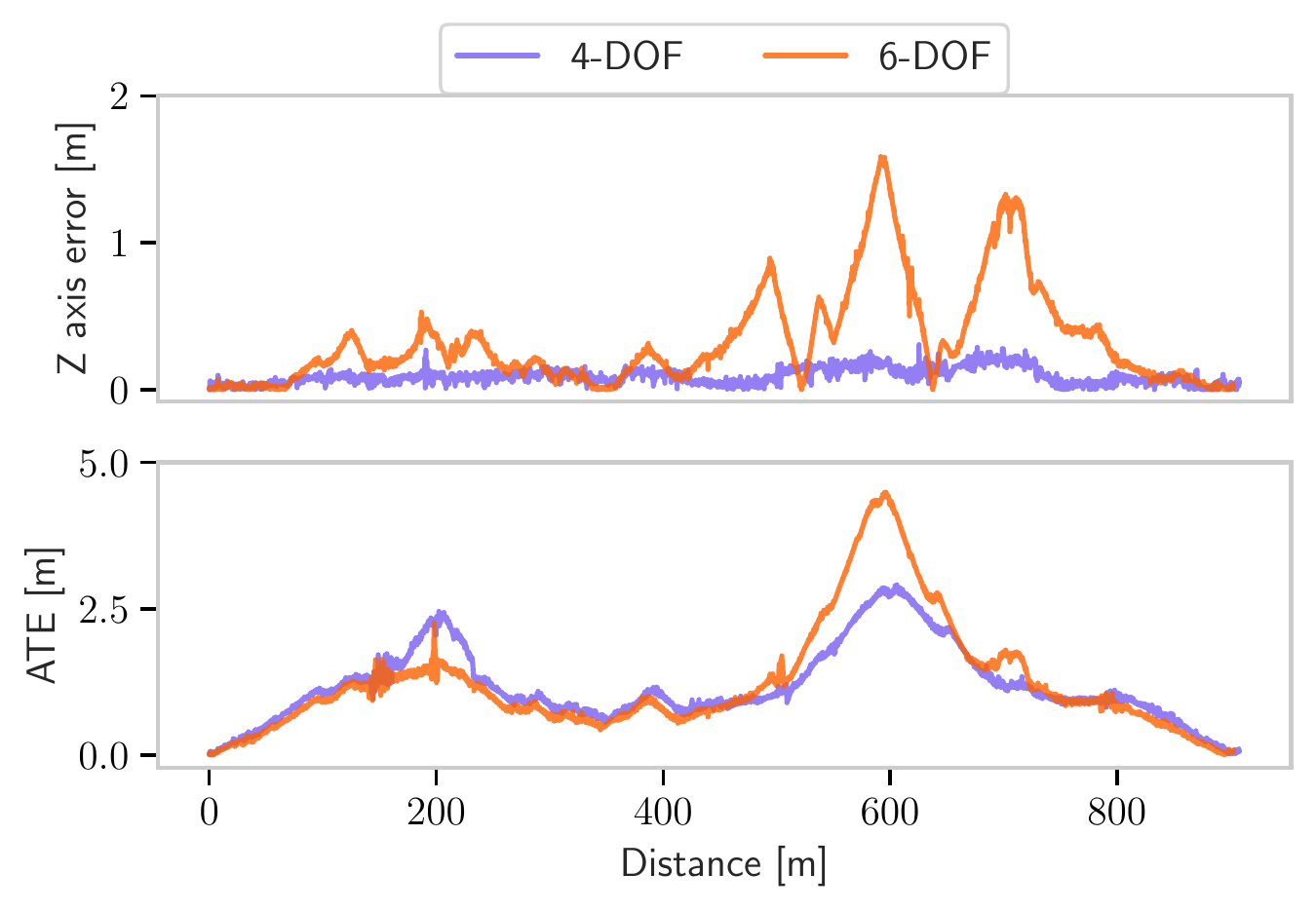}
    \caption{Comparison of the 4-DOF and 6-DOF ATE during the Mine and Cave experiment in the underground environment. The 4-DOF suppresses the vertical drift (top) which improves the overall ATE (bottom).
    }
    \label{fig:darpa_cave_error}
\end{figure}

\begin{figure}[htbp]
    \centering
    \includegraphics[trim=0 0 0 0,clip, width=\linewidth]{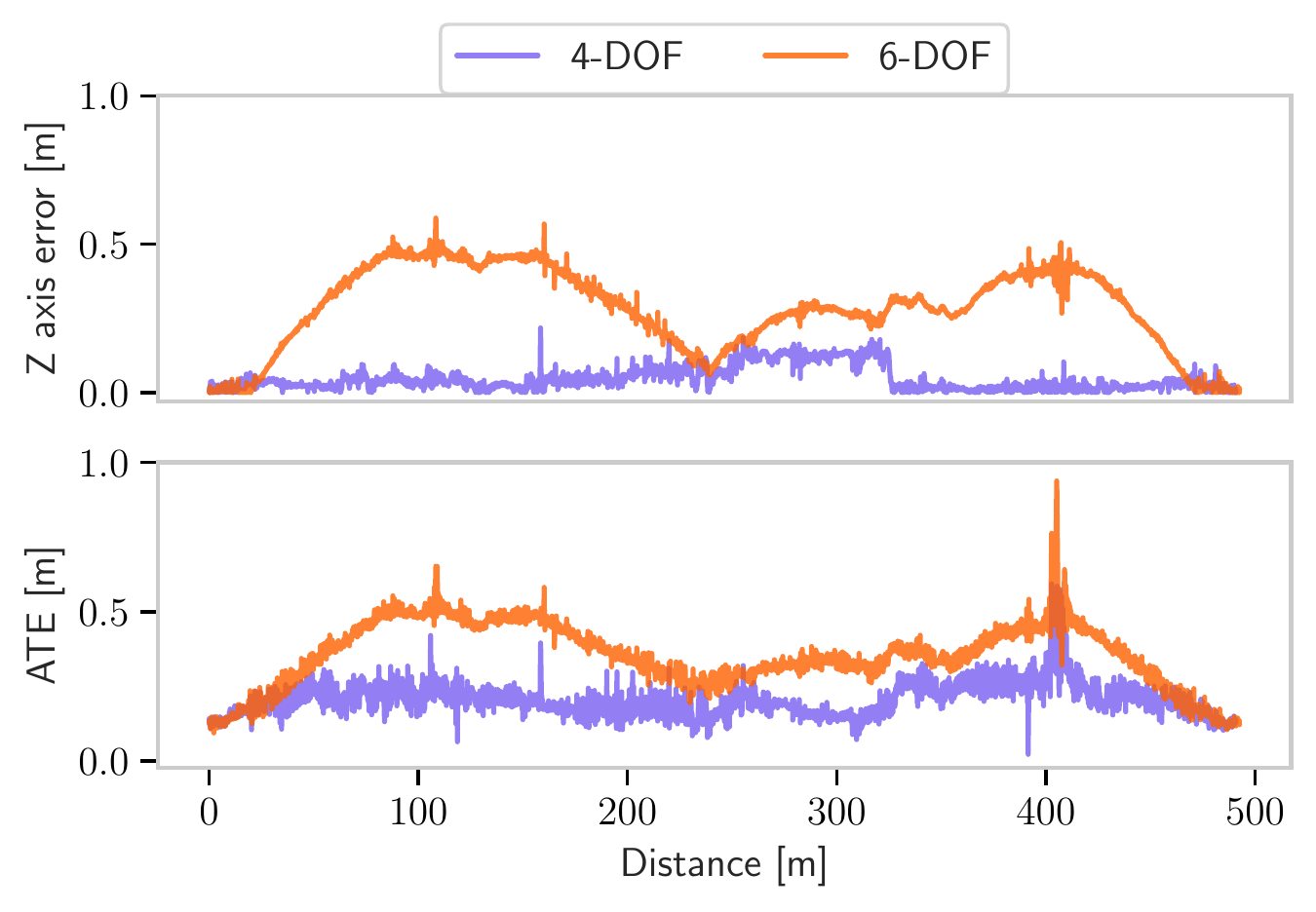}
    \caption{ATE in the Urban part of the underground environment. The vertical drift was  the dominant contributor to ATE in this experiment.
    }
    \label{fig:darpa_urban_error}
\end{figure}

Our general experience from the underground environment is that it is well suited for the point-to-plane \ac{ICP} error function.
In the case of the Mine and Cave experiment, the maximum ATE reaches only three to four meters on a trajectory \SI{950}{\m} long, depending on the 4-DOF or classical 6-DOF \ac{ICP} choice (\autoref{fig:darpa_cave_error}).
The shorter Urban experiment performs similarly well, less than one meter in \SI{450}{\m} (\autoref{fig:darpa_urban_error}).
We attribute this to the modern 360-degree lidars and plentiful surfaces in this particular underground environment.
Also, the course did not contain long uniform tunnels, which are challenging for this type of \ac{SLAM}.
Nevertheless, we observe a trend when switching between 6-DOF and 4-DOF localization: the drift in the \textbf{z} coordinate is suppressed in both experiments, see \autoref{fig:darpa_cave_error} and \autoref{fig:darpa_urban_error}.
Both approaches drift on the \textbf{x}-\textbf{y} plane, but this is expected since the \ac{IMU} does not add any constraints in this plane.
In the Mine and Cave experiment, the ATE of the 4-DOF variant (see \autoref{fig:darpa_cave_error}) momentarily overshoots the 6-DOF error at around \SI{200}{\m} distance due to drift in the \textbf{x}-\textbf{y} plane, but drift in \textbf{z} eventually dominates when the robot reaches the edge of the competition course.
In the Urban experiment, the \textbf{z} drift is the dominant contributor to ATE along the whole trajectory and thus the 4-DOF variant helps by suppressing it below \SI{25}{\cm} in the worst case.
In both experiments, the ATE drops back to zero as the robot returns to the starting point.

\subsection{Montmorency forest large-scale outdoor environment}

\begin{figure}[htbp]
    \centering
    \includegraphics[trim=0 0 0 0,clip, width=\linewidth]{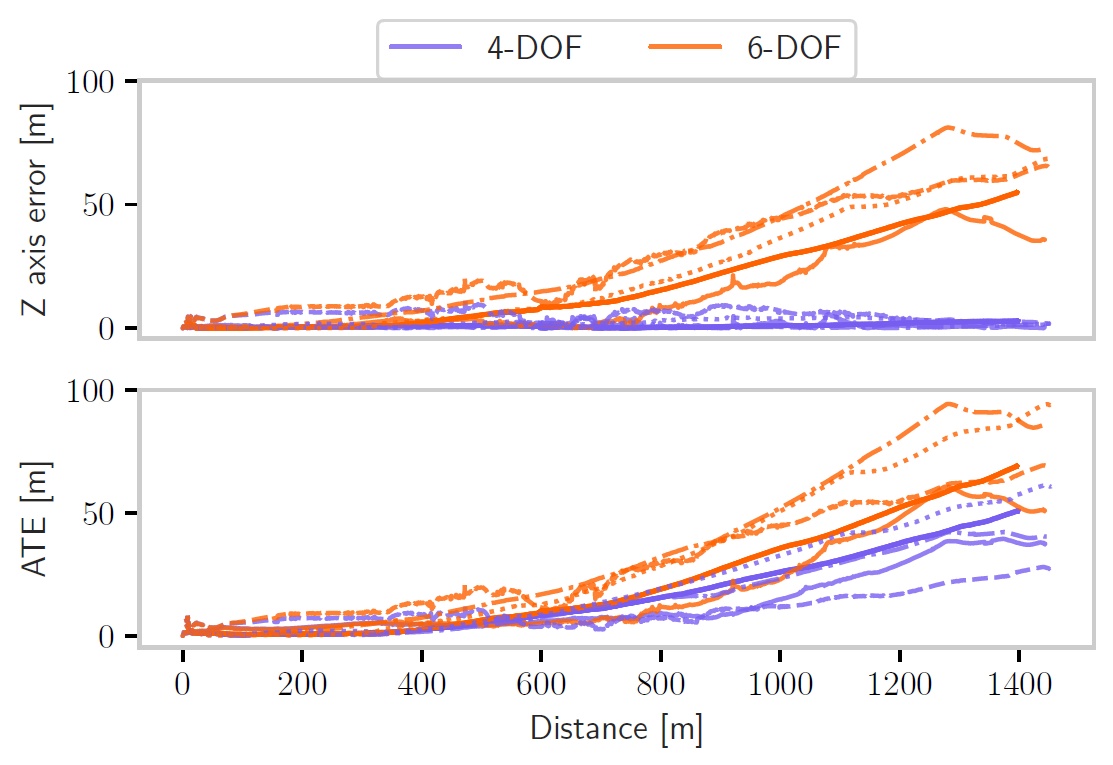}
    \caption{ATE of the Montmorency forest experiments. Five runs shown together, each with its unique line style. The effect of 4-DOF is evident in the Z axis error.
    }
    \label{fig:forest_error}
\end{figure}

\autoref{fig:drift} demonstrates the difference between the 4-DOF represented with a green map and the 6-DOF represented in red.
Towards the end of the \SI{1.5}{\km} trail, the elevation error can reach \SI{55}{\m}.
Unconstrained, the map drifts mainly in the narrow forest corridors, possibly due to the fuzzy vegetation geometry. 
Incorporating the gravity information in the case of the 4-DOF \ac{ICP} suppresses this effect.
Figure \ref{fig:forest_error} assembles all five runs in one plot, each run with its unique line style.
Although the error differs between each run and two runs went in the opposite direction, the trend is clear.
Both 4-DOF and 6-DOF drift in the \textbf{x}-\textbf{y} plane, but the 4-DOF always limits the \textbf{z} drift.
This translates to the total ATE shown in \autoref{fig:forest_error} with the slower 4-DOF localization error growth compared to the counterparts in 6-DOF. 
%
%

\subsection{Normalized ATE analysis}

\begin{figure}[htbp]
    \centering
    \includegraphics[trim=5 0 0 0,clip, width=\linewidth]{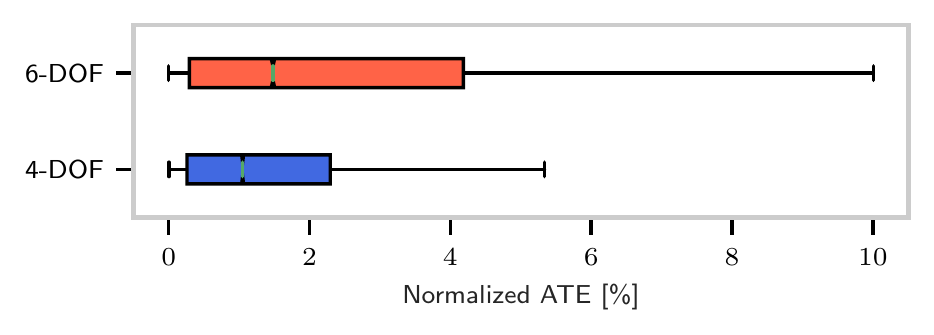}
    \caption{Normalized ATE of the 4-DOF and 6-DOF variants over all poses and experiments. The first quartile, median and third quartile values are $(0.26|1.05|2.29)$\% and $(0.3|1.48|4.19)$\%, respectively. The black whiskers extend to datapoints in the 1.5x inter-quartile range.
    }
    \label{fig:normalized_ate}
\end{figure}

To examine what is the expected localization drift with 4-DOF and the standard 6-DOF ICP variants, we have assembled ATE from all poses and experiments and normalized it by the distance traveled.
\autoref{fig:normalized_ate} shows the resulting distributions. 
4-DOF limits the median value from \SI{1.48}{\%} to \SI{1.05}{\%} and also the upper half of the distribution gets compressed with the new third quartile \SI{2.29}{\%} instead of \SI{4.19}{\%}.
This result shows that we cannot eliminate the drift, but we can significantly reduce it by exploiting the \ac{IMU} modality, which is usually available on robotic platforms anyway.

\subsection{Practical considerations}

The presented benefits of the 4-DOF modification are conditioned by correctly estimated roll and pitch angles, and by precise \ac{IMU}-\ac{lidar} coordinate frame calibration.
In our experiments, we use the Xsens MTI-30 and MTI-100 \acp{IMU} which provide sub-degree orientation accuracy in the roll and pitch angles.
More importantly, imprecise alignment between the \ac{IMU} and \ac{lidar} frames can lead to worse drift than with the classical 6-DOF localization.
Our practical approach to the calibration is to use a sufficiently large, flat floor and to adjust the transformation between the sensor frames until the captured point cloud appears flat when transformed to the inertial coordinate frame.
The calibration can be further fine-tuned by performing a mapping run in a sufficiently large loop (i.e., more than \SI{500}{\m}) and observing the resulting elevation mismatch.
The tangent between the loop length and the error elevation is the final correction of the lidar coordinate frame pitch angle in the inertial frame.
This correction can be as low as \SI{0.1}{\degree} and is sufficient to achieve the results presented in this paper.

\section{Conclusion}
\label{sec:conclusion}
In this paper, we have proposed a tightly-coupled variant of the \ac{ICP} algorithm, which exploits directly observable \acp{DOF} in the \ac{IMU} orientation.
It reduces the median normalized ATE by \SI{30}{\%} compared to the standard 6-DOF \ac{ICP} by limiting the vertical drift.
The method requires accurate \ac{IMU} orientation estimates and \ac{IMU}-\ac{lidar} calibration.
This condition is however achievable with standard sensors and tools.
Moreover, the code is readily available to the community as a part of the \texttt{libpointmatcher} library.
We believe it can be useful as a module in complex graph-based SLAM systems as well as a lightweight lidar odometry with reduced vertical drift.

The 4-DOF \ac{ICP} has been derived for the point-to-plane error function and implicitly supports the point-to-gaussian error function.
Yet, the family of \ac{ICP} variants is vast and it is for future work to extend this idea to more of them.
Moreover, it would be practical to replace the manual \ac{IMU}-\ac{lidar} calibration be an automatic one based on additional external measurements, such as GPS or global map optimization.


\section*{Acknowledgment}

The trajectory analysis in this paper was done using \emph{Python package for the evaluation of odometry and SLAM} available at \url{https://github.com/MichaelGrupp/evo}.

\addtolength{\textheight}{-2cm}   

\printbibliography

@InProceedings{Landry2019,
  author    = {Landry, David and Pomerleau, Francois and Giguère, Philippe},
  title     = {{CELLO-3D}: Estimating the Covariance of ICP in the Real World},
  booktitle = {2019 International Conference on Robotics and Automation (ICRA)},
  year      = {2019},
  pages     = {8190-8196},
  doi       = {10.1109/ICRA.2019.8793516},
}

@article{Thrun2006,
author = {Thrun, S. and Montemerlo, M.},
journal = {IJRR},
number = {5-6},
pages = {403--429},
title = {{The GraphSLAM Algorithm with Applications to Large-Scale Mapping of Urban Structures}},
volume = {25},
year = {2006}
}

@article{Cadena2016,
   author = {Cesar Cadena and Luca Carlone and Henry Carrillo and Yasir Latif and Davide Scaramuzza and Jose Neira and Ian Reid and John J. Leonard},
   doi = {10.1109/TRO.2016.2624754},
   issn = {1552-3098},
   issue = {6},
   journal = {IEEE Transactions on Robotics},
   month = {12},
   pages = {1309-1332},
   title = {Past, Present, and Future of Simultaneous Localization and Mapping: Toward the Robust-Perception Age},
   volume = {32},
   year = {2016},
}

@article{Zlot2014,
   author = {Robert Zlot and Michael Bosse},
   doi = {10.1002/rob.21504},
   issn = {15564959},
   issue = {5},
   journal = {Journal of Field Robotics},
   month = {9},
   pages = {758-779},
   title = {Efficient Large-scale Three-dimensional Mobile Mapping for Underground Mines},
   volume = {31},
   year = {2014},
}

@article{Palieri2021,
   author = {Matteo Palieri and Benjamin Morrell and Abhishek Thakur and Kamak Ebadi and Jeremy Nash and Arghya Chatterjee and Christoforos Kanellakis and Luca Carlone and Cataldo Guaragnella and Ali Akbar Agha-Mohammadi},
   doi = {10.1109/LRA.2020.3044864},
   issn = {23773766},
   issue = {2},
   journal = {IEEE Robotics and Automation Letters},
   title = {LOCUS: A Multi-Sensor Lidar-Centric Solution for High-Precision Odometry and 3D Mapping in Real-Time},
   volume = {6},
   year = {2021},
}

@InProceedings{Shan2020,
  author    = {Tixiao Shan and Brendan Englot and Drew Meyers and Wei Wang and Carlo Ratti and Daniela Rus},
  title     = {{LIO-SAM}: Tightly-coupled lidar inertial odometry via smoothing and mapping},
  booktitle = {IEEE International Conference on Intelligent Robots and Systems},
  year      = {2020},
  doi       = {10.1109/IROS45743.2020.9341176},
  issn      = {21530866},
}

@InProceedings{Ye2019,
  author    = {Haoyang Ye and Yuying Chen and Ming Liu},
  title     = {Tightly coupled 3D Lidar inertial odometry and mapping},
  booktitle = {IEEE International Conference on Robotics and Automation},
  year      = {2019},
  volume    = {2019-May},
  doi       = {10.1109/ICRA.2019.8793511},
  issn      = {10504729},
}

@article{Forster2017,
   author = {Christian Forster and Luca Carlone and Frank Dellaert and Davide Scaramuzza},
   doi = {10.1109/TRO.2016.2597321},
   issn = {15523098},
   issue = {1},
   journal = {IEEE Transactions on Robotics},
   title = {On-Manifold Preintegration for Real-Time Visual-Inertial Odometry},
   volume = {33},
   year = {2017},
}

@article{Zhang2018,
   author = {Zichao Zhang and Guillermo Gallego and Davide Scaramuzza},
   doi = {10.1109/LRA.2018.2833152},
   issn = {23773766},
   issue = {3},
   journal = {IEEE Robotics and Automation Letters},
   title = {On the Comparison of Gauge Freedom Handling in Optimization-Based Visual-Inertial State Estimation},
   volume = {3},
   year = {2018},
}

@Misc{Agha2021nebula,
  author        = {Ali Agha and Kyohei Otsu and Benjamin Morrell and David D. Fan and Rohan Thakker and Angel Santamaria-Navarro and Sung-Kyun Kim and Amanda Bouman and Xianmei Lei and Jeffrey Edlund and Muhammad Fadhil Ginting and Kamak Ebadi and Matthew Anderson and Torkom Pailevanian and Edward Terry and Michael Wolf and Andrea Tagliabue and Tiago Stegun Vaquero and Matteo Palieri and Scott Tepsuporn and Yun Chang and Arash Kalantari and Fernando Chavez and Brett Lopez and Nobuhiro Funabiki and Gregory Miles and Thomas Touma and Alessandro Buscicchio and Jesus Tordesillas and Nikhilesh Alatur and Jeremy Nash and William Walsh and Sunggoo Jung and Hanseob Lee and Christoforos Kanellakis and John Mayo and Scott Harper and Marcel Kaufmann and Anushri Dixit and Gustavo Correa and Carlyn Lee and Jay Gao and Gene Merewether and Jairo Maldonado-Contreras and Gautam Salhotra and Maira Saboia Da Silva and Benjamin Ramtoula and Yuki Kubo and Seyed Fakoorian and Alexander Hatteland and Taeyeon Kim and Tara Bartlett and Alex Stephens and Leon Kim and Chuck Bergh and Eric Heiden and Thomas Lew and Abhishek Cauligi and Tristan Heywood and Andrew Kramer and Henry A. Leopold and Chris Choi and Shreyansh Daftry and Olivier Toupet and Inhwan Wee and Abhishek Thakur and Micah Feras and Giovanni Beltrame and George Nikolakopoulos and David Shim and Luca Carlone and Joel Burdick},
  title         = {NeBula: Quest for Robotic Autonomy in Challenging Environments; TEAM CoSTAR at the DARPA Subterranean Challenge},
  howpublished  = {arXiv preprint arXiv:2103.11470},
  year          = {2021},
  archiveprefix = {arXiv},
  eprint        = {2103.11470},
  primaryclass  = {cs.RO},
}

@article{Kelly2011,
   author = {Jonathan Kelly and Gaurav S. Sukhatme},
   doi = {10.1177/0278364910382802},
   issn = {17413176},
   issue = {1},
   journal = {International Journal of Robotics Research},
   title = {Visual-inertial sensor fusion: Localization, mapping and sensor-to-sensor Self-calibration},
   volume = {30},
   year = {2011},
}

@article{Jones2011,
   author = {Eagle S. Jones and Stefano Soatto},
   doi = {10.1177/0278364910388963},
   issn = {02783649},
   issue = {4},
   journal = {International Journal of Robotics Research},
   title = {Visual-inertial navigation, mapping and localization: A scalable real-time causal approach},
   volume = {30},
   year = {2011},
}

@article{Lupton2012,
   author = {Todd Lupton and Salah Sukkarieh},
   doi = {10.1109/TRO.2011.2170332},
   issn = {15523098},
   issue = {1},
   journal = {IEEE Transactions on Robotics},
   title = {Visual-inertial-aided navigation for high-dynamic motion in built environments without initial conditions},
   volume = {28},
   year = {2012},
}

@InProceedings{Kaess2011,
  author    = {Michael Kaess and Hordur Johannsson and Richard Roberts and Viorela Ila and John Leonard and Frank Dellaert},
  title     = {iSAM2: Incremental smoothing and mapping with fluid relinearization and incremental variable reordering},
  booktitle = {2011 IEEE International Conference on Robotics and Automation},
  year      = {2011},
  pages     = {3281-3288},
  month     = {5},
  publisher = {IEEE},
  doi       = {10.1109/ICRA.2011.5979641},
  isbn      = {978-1-61284-386-5},
  journal   = {2011 IEEE International Conference on Robotics and Automation},
}

@InProceedings{Nguyen2021,
  author    = {Nguyen, Thien-Minh and Cao, Muqing and Yuan, Shenghai and Lyu, Yang and Nguyen, Thien Hoang and Xie, Lihua},
  title     = {LIRO: Tightly Coupled Lidar-Inertia-Ranging Odometry},
  booktitle = {2021 IEEE International Conference on Robotics and Automation (ICRA)},
  year      = {2021},
  pages     = {14484-14490},
  doi       = {10.1109/ICRA48506.2021.9560954},
}

@InProceedings{Qin2020,
  author    = {Chao Qin and Haoyang Ye and Christian E. Pranata and Jun Han and Shuyang Zhang and Ming Liu},
  title     = {{LINS}: A Lidar-Inertial State Estimator for Robust and Efficient Navigation},
  booktitle = {IEEE International Conference on Robotics and Automation},
  year      = {2020},
  doi       = {10.1109/ICRA40945.2020.9197567},
  issn      = {10504729},
}

@article{Besl1992,
author = {Besl, P. J. and McKay, N. D.},
journal = {TPAMI},
number = {2},
pages = {239--256},
title = {{A method for registration of 3-D shapes}},
volume = {14},
year = {1992}
}

@article{Pomerleau2013,
author = {Pomerleau, F. and Colas, F. and Siegwart, R. and Magnenat, S.},
journal = {Autonomous Robots},
number = {3},
pages = {133--148},
title = {{Comparing ICP variants on real-world data sets: Open-source library and experimental protocol}},
volume = {34},
year = {2013}
}

@inproceedings{Babin2019a,
  title = {Large-scale 3{D} Mapping of Subarctic Forests},
  booktitle = {Proceedings of the Conference on Field and Service Robotics (FSR). Springer Tracts in Advanced Robotics},
  year = {2019},
  author = {Babin, P. and Dandurand, P. and Kubelka, V. and Gigu{\`e}re, P. and Pomerleau, F.},
  project = {penalty_icp,libpointmatcher},
}

@Article{Chen1992,
  author   = {Yang Chen and Gérard Medioni},
  title    = {Object modelling by registration of multiple range images},
  journal  = {Image and Vision Computing},
  year     = {1992},
  volume   = {10},
  number   = {3},
  pages    = {145-155},
  issn     = {0262-8856},
  note     = {Range Image Understanding},
  doi      = {https://doi.org/10.1016/0262-8856(92)90066-C},
  url      = {https://www.sciencedirect.com/science/article/pii/026288569290066C},
}

@article{Pomerleau2015a,
  year = {2015},
  publisher = {Now Publishers},
  volume = {4},
  number = {1},
  pages = {1--104},
  author = {Pomerleau, F. and Colas, F. and Siegwart, R.},
  title = {A Review of Point Cloud Registration Algorithms for Mobile Robotics},
  journal = {Foundations and Trends in Robotics},
  project = {libpointmatcher}
}

@misc{Baril2021,
  title = {Kilometer-scale autonomous navigation in subarctic forests: challenges and lessons learned},
  howpublished = {arXiv preprint arXiv:2111.13981},
  year = {2021},
  author = {Baril, D. and Deschênes, S. and Gamache, O. and Vaidis, M. and LaRocque, D. and Laconte, J. and Kubelka, V. and Giguère, P. and Pomerleau, F.},
  project = {snow}
}

@InProceedings{Sturm2012,
  author    = {Jrgen Sturm and Nikolas Engelhard and Felix Endres and Wolfram Burgard and Daniel Cremers},
  title     = {A benchmark for the evaluation of {RGB-D} {SLAM} systems},
  booktitle = {IEEE International Conference on Intelligent Robots and Systems},
  year      = {2012},
  doi       = {10.1109/IROS.2012.6385773},
  issn      = {21530858},
}

\end{document}